\pdfoutput=1

\documentclass[11pt]{article}

\usepackage{acl}

\usepackage{times}
\usepackage{latexsym}

\usepackage{booktabs} 
\usepackage{tabularx} 
\usepackage{multirow}
\usepackage{graphicx}
\usepackage{xcolor}
\usepackage[normalem]{ulem} 
\newcommand\hlgreen{\bgroup\markoverwith
  {\textcolor{green}{\rule[-.5ex]{2pt}{2.5ex}}}\ULon}
\newcommand\hlorange{\bgroup\markoverwith
  {\textcolor{orange}{\rule[-.5ex]{2pt}{2.5ex}}}\ULon}
\usepackage{scalerel,xparse}

\NewDocumentCommand\emojismiley{}{
    \includegraphics[scale=0.1]{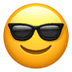}
}

\usepackage[T1]{fontenc}

\usepackage{kotex}
\usepackage{CJKutf8}
\usepackage{hyperref}

\usepackage[utf8]{inputenc}

\usepackage{enumitem}

\usepackage{microtype}
\usepackage{url}

\usepackage{graphicx}

\usepackage{multirow}

\title{\includegraphics[height=3.6mm]{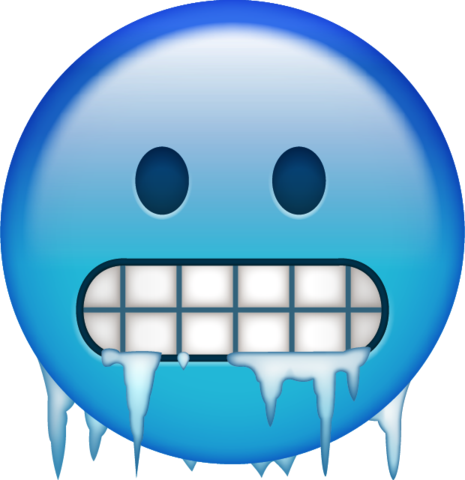} KOLD: Korean Offensive Language Dataset}
\author
{
Younghoon Jeong\textsuperscript{\dag} \qquad Juhyun Oh\qquad  Jongwon Lee\\
    \textbf{Jaimeen Ahn\textsuperscript{\dag} \qquad Jihyung Moon\textsuperscript{\ddag} \qquad Sungjoon Park\textsuperscript{\ddag} \qquad Alice Oh\textsuperscript{\dag}} \vspace{3mm} \\
    \textsuperscript{\dag}School of Computing, KAIST \vspace{1mm} \\
    \textsuperscript{\ddag}SoftlyAI Research, SoftlyAI 
    \vspace{3mm} \\
    \texttt{\small hoon2j@gmail.com},\ \texttt{\small 411juhyun@snu.ac.kr}, \
    \texttt{\small mybizzer@gmail.com},  \\
    \texttt{\small jaimeen01@kaist.ac.kr},\ 
    \texttt{\small \{jihyung.moon, sungjoon.park\}@softly.ai}, \\
    \texttt{\small alice.oh@kaist.edu} \\
}

\begin{document}
\maketitle
\begin{abstract}
\textit{\textbf{Warning}: this paper contains content that may be offensive or upsetting.}

Recent directions for offensive language detection are \textit{hierarchical modeling}, identifying the type and the target of offensive language, and \textit{interpretability} with offensive span annotation and prediction.

These improvements are focused on English and do not transfer well to other languages because of cultural and linguistic differences.

In this paper, we present the Korean Offensive Language Dataset (KOLD) comprising 40,429 comments, which are annotated hierarchically with the type and the target of offensive language, accompanied by annotations of the corresponding text spans. We collect the comments from NAVER 
news and YouTube platform and provide the titles of the articles and videos as the context information for the annotation process.

We use these annotated comments as training data for Korean BERT and RoBERTa models and find that they are effective at offensiveness detection, target classification, and target span detection while having room for improvement for target group classification and offensive span detection. We discover that the target group distribution differs drastically from the existing English datasets, and observe that providing the context information improves the model performance in offensiveness detection (+0.3), target classification (+1.5), and target group classification (+13.1).
We publicly release the dataset and baseline models.
\end{abstract}

\section{Introduction}
\begin{table*}[!htp]
\centering
\resizebox{\textwidth}{!}{%
\begin{tabular}{@{}llcccc@{}}
\toprule
\multicolumn{1}{c}{}                                                   & \multicolumn{1}{c}{}                                   & \multicolumn{4}{c}{\textbf{Label}}                                                                                            \\ \cmidrule{3-6} 
\multicolumn{1}{c}{}                                                   & \multicolumn{1}{c}{}                                   &                                          &                                        & \multicolumn{2}{c}{\textbf{Level C}} \\ \cmidrule{5-6}
\multicolumn{1}{c}{\multirow{-3}{*}{\textbf{Context (title)}}}                   & \multicolumn{1}{c}{\multirow{-3}{*}{\textbf{Comment}}} &
\multirow{-2}{*}{\textbf{Level A}} & \multirow{-2}{*}{\textbf{Level B}} & \textbf{\begin{tabular}[c]{@{}c@{}}Target \\ Group \\ Attribute \end{tabular}} & \textbf{\begin{tabular}[c]{@{}c@{}}Target \\ Group \end{tabular}} \\ \midrule
\begin{tabular}[c]{@{}l@{}}심화하는 젠더갈등...SNS 때문? \\ Gender in Conflict, deepening by SNS?\end{tabular} &
\begin{tabular}[c]{@{}l@{}}사람은 사람일뿐 \\ People are all just people\end{tabular}  &
  NOT \\ \midrule
\begin{tabular}[c]{@{}l@{}}한국 106개 사회단체, `아프가니스탄 난민 보호책' 수립 촉구 \\ 'Protection of Afghan Refugees' by 106 Korean civil societies\end{tabular} &
\begin{tabular}[c]{@{}l@{}}\colorbox[HTML]{C1D6EB}{정신좀 차리자} \\ \colorbox[HTML]{C1D6EB}{Snap out of it}\end{tabular}   &
  OFF &
  UNT \\ \midrule
\begin{tabular}[c]{@{}l@{}}북한, 아프간 사태로 자신감 얻었나... 미국에 `인권 문제' 역공 \\ N. Korea calls US-led Afghan war 'Human Rights Crime'\end{tabular} &
  \begin{tabular}[c]{@{}l@{}} 누가 \colorbox[HTML]{C9DFB8}{김정은} \colorbox[HTML]{C1D6EB}{인권 좀 말살 안하나?} \\ Anyone for \colorbox[HTML]{C1D6EB}{the extermination of} \colorbox[HTML]{C9DFB8}{Kim Jong-un}? \end{tabular} &
  OFF &
  IND \\ \midrule
\begin{tabular}[c]{@{}l@{}}틱톡 감성 거부하다 못해 패버리는 흑인 형님 모음 \\ Angry TikTok reactions by Black bro\end{tabular} &
\begin{tabular}[c]{@{}l@{}}\colorbox[HTML]{C9DFB8}{틱톡} \colorbox[HTML]{C1D6EB}{이라는 이름의 쓰래기는 제거되야되} \\ \colorbox[HTML]{C1D6EB}{The garbage named} \colorbox[HTML]{C9DFB8}{TikTok} \colorbox[HTML]{C1D6EB}{should be eliminated} \end{tabular} &
  OFF &
  OTH \\ \midrule
\begin{tabular}[c]{@{}l@{}}"이 법은 죽을 뻔한 누구를 살리기도 해요" \\ "This law could save desperate lives who almost died"\end{tabular} &
  \begin{tabular}[c]{@{}l@{}}수술받은 \colorbox[HTML]{C9DFB8}{트랜스젠더들}을 봐도 \\ \colorbox[HTML]{C1D6EB}{분노가 솟아 칼을 꺼내 죽이겠다!} \\ \colorbox[HTML]{C1D6EB}{I get mad} when I see \colorbox[HTML]{C9DFB8}{transgender people} who had \\ surgery,  \colorbox[HTML]{C1D6EB}{I'll take a knife and kill them all!} \end{tabular} &
  OFF &
  GRP &
  \begin{tabular}[c]{@{}c@{}}Gender \& \\ Sexual \\ Orientation \end{tabular} &
  LGBTQ+ \\ \midrule

\begin{tabular}[c]{@{}l@{}}치명적인 유혹 - 조선족이랑 결혼한 후기 \\ Deadly Temptation - A marriage with Korean-Chinese\end{tabular} &
  \begin{tabular}[c]{@{}l@{}}\colorbox[HTML]{C9DFB8}{조선족}들 \colorbox[HTML]{C1D6EB}{중국으로 다돌아갔으면 속이 시원하겠다\emojismiley} \\ \colorbox[HTML]{C1D6EB}{It would be so refreshing} if all the \colorbox[HTML]{C9DFB8}{Korean-Chinese} \\ \colorbox[HTML]{C1D6EB}{went back to China\emojismiley}\end{tabular} &
  OFF &
  GRP &
  \begin{tabular}[c]{@{}c@{}}Race, \\ Ethnicity \& \\ Nationality\end{tabular} &
  \begin{tabular}[c]{@{}c@{}}Korean-\\ Chinese\end{tabular} \\ \midrule
  
\begin{tabular}[c]{@{}l@{}}한국의 “이슬람교” \\ ``Islam'' in Korea \end{tabular} &
\begin{tabular}[c]{@{}l@{}}뭘 믿든상관 없는데 \colorbox[HTML]{C1D6EB}{사람을 죽이니까 무서운거지} \\ I don’t care what (they) believe, \colorbox[HTML]{C1D6EB}{what matters is the fact} \\ \colorbox[HTML]{C1D6EB}{that (they) kill people} \end{tabular}&
  OFF &
  GRP &
  \begin{tabular}[c]{@{}c@{}}Race, \\ Ethnicity \& \\ Nationality\end{tabular} & Muslim \\ \midrule

\begin{tabular}[c]{@{}l@{}}GS25, 브레이브걸스 포스터 또 젠더 이슈 논란 \\ GS25's poster of Brave Girls evokes dispute over gender \\ issues yet again\end{tabular} &
\begin{tabular}[c]{@{}l@{}}\colorbox[HTML]{C9DFB8}{페미}는 젠더가 아니라 \colorbox[HTML]{C1D6EB}{정신병이다} \\ \colorbox[HTML]{C9DFB8}{Feminism} is not about gender inequality, it is a \colorbox[HTML]{C1D6EB}{mental} \\ \colorbox[HTML]{C1D6EB}{disorder} \end{tabular}&
  OFF &
  GRP &
  Misc. &
  Feminists \\ \bottomrule

\end{tabular}%
}
\caption{
Examples of the comments in KOLD, along with the annotation results. The \textit{Title} is either the headline of the news article or the title of the video where the comment is posted on. The subject in the parentheses is omitted in the original sentence in Korean. (OFF: offensive, NOT: not offensive, UNT: untargeted, IND: individual, OTH: other, GRP: group, \colorbox[HTML]{C1D6EB}{blue}: offensive span, \colorbox[HTML]{C9DFB8}{green}: target span)
}
\label{tab:my-table}
\end{table*}

\begin{figure}[t!]
    \centering
    \includegraphics[width=\columnwidth]{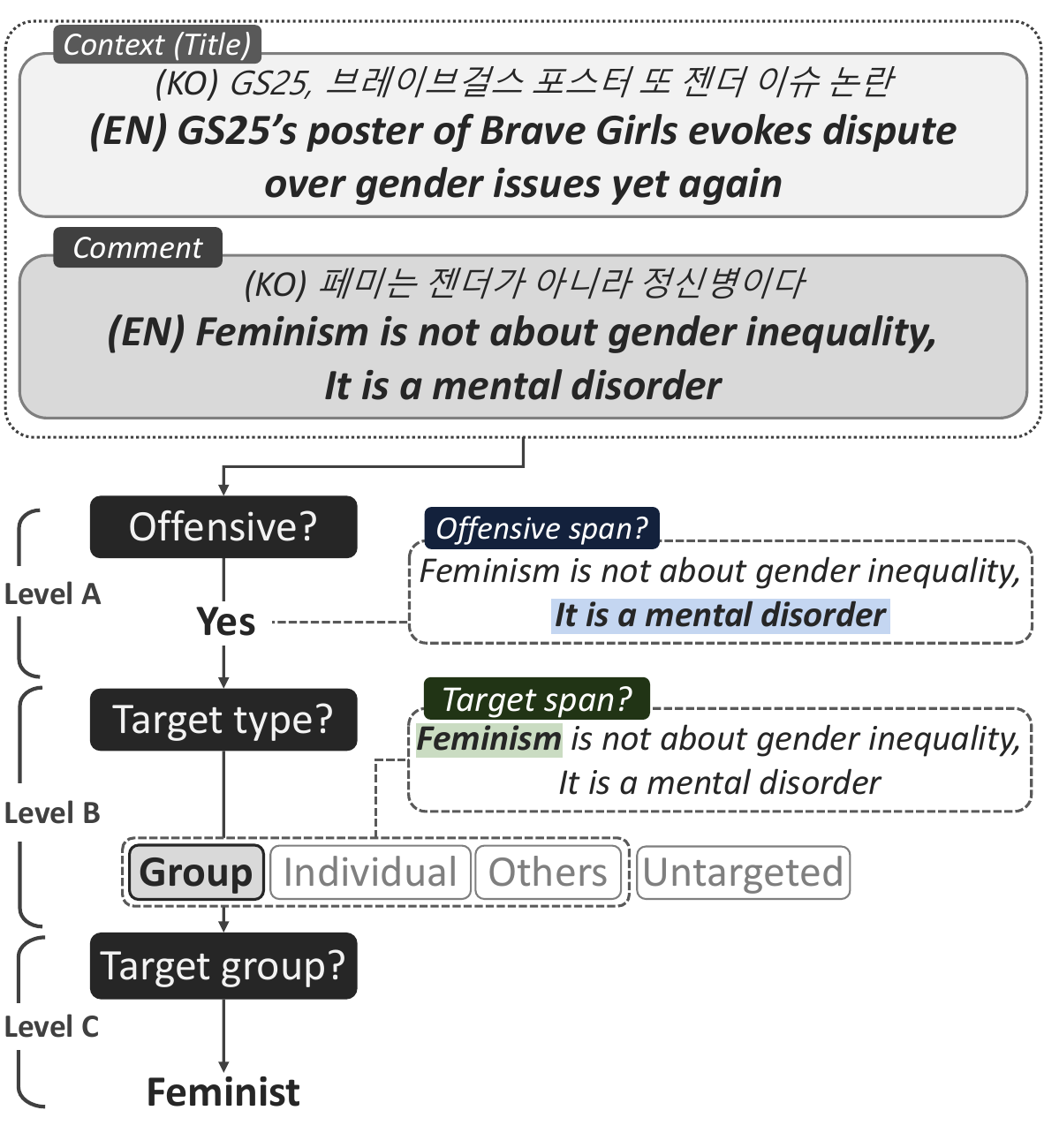}
    \caption{An illustration of the annotation process of KOLD. The title is given to provide context to the annotators. Along with the categorical labels, annotators are asked to mark spans of a sentence that justifies their decision.}
    \label{fig:example}
\end{figure}

Online offensive language is a growing societal problem. It propagates negative stereotypes about the targeted social groups, causing representational harm \citep{barocas2017problem}.
Among various research directions for offensive language detection,
\citet{waseem2017understanding} proposes a taxonomy to distinguish hate speech and cyberbullying based on whether the offensive language is directed toward a specific individual or entity, or toward a generalized group.
\citet{zampieri-etal-2019-predicting} integrates this taxonomy into a hierarchical annotation process to create the Offensive Language Identification Dataset (OLID), which is adopted in other languages as well \citep{zeinert-etal-2021-annotating, zampieri-etal-2020-semeval}.

Many problems remain in offensive language detection such as failure to generalize \citep{grondahl2018all, karan-snajder-2018-cross}, over-sensitivity to commonly-attacked identities \citep{dixon2018measuring, kennedy-etal-2020-contextualizing}, and propagating bias in annotation \citep{sap-etal-2019-risk, davidson-etal-2019-racial}. Among various attempts to solve those problems, first is offensive language target detection to identify the individuals or groups who are the victims of offensive language \citep{sap-etal-2020-social, mathew2020hatexplain, shvets-etal-2021-targets}. Second is offensive language span detection to provide some explanation for offensive language \citep{pavlopoulos-etal-2022-detection, mathew2020hatexplain}. Third is to provide context information of offensive language to improve offensive language detection \citep{vidgen-etal-2021-introducing,de-gibert-etal-2018-hate,gao-huang-2017-detecting}.

However, these steps forward are limited to English because we lack comprehensive datasets in other languages \citep{mubarak-etal-2017-abusive, fortuna-etal-2019-hierarchically, chiril-etal-2020-annotated, rizwan-etal-2020-hate}.
Language-specific datasets are essential for offensive language which is very culture- and language-dependent \cite{reichelmann-2021-hate-knows, albadi2018they, ousidhoum-etal-2019-multilingual}.
Although Korean is a comparably high-resourced language \cite{joshi-etal-2020-state}, there are only a few publicly available Korean offensive language corpora \citep{moon-etal-2020-beep}, and they do not consider the type of the target (e.g., group, individual) to differentiate hate speech from cyberbullying, include context information such as titles of articles, nor annotate text spans for explainability.

In this paper, we describe and publicly release the Korean Offensive Language Dataset (KOLD), 40,429 comments collected from news articles and videos.\footnote{KOLD: Korean Offensive Language Dataset is available at \href{https://github.com/boychaboy/KOLD}{http://github.com/boychaboy/KOLD}.}
The unique characteristics of our dataset are as follows:

\begin{itemize}
    \item It is the first Korean dataset with a hierarchical taxonomy of offensive language (see Figure \ref{fig:example}). If the comment is group-targeted offensive language, we additionally annotate among the 21 target group labels tailored to Korean culture.
    \item The specific spans of text that are offensive or that reveal targeted communities are annotated (see Table \ref{tab:my-table} for examples). KOLD is the first publicly released dataset to provide both types of spans for offensive language in Korean.
    \item The comments in our dataset are annotated with the original context. We provide the titles of the articles and videos during the annotation process, which resembles the realistic setting of actual usage. 
\end{itemize}

\section{Annotation Task Design}
We use a hierarchical annotation framework based on the multi-layer annotation schema in OLID \citep{zampieri-etal-2019-predicting}. Additionally, we identify the specific target group of the offensive language.
We also annotate the spans that support the labeling decision if the comment is offensive and/or contains a target of offensiveness.
Figure~\ref{fig:example} illustrates an overview of our annotation task, and Table~\ref{tab:my-table} shows examples. 

\subsection{Level A: Offensive Language Detection}
\label{sec:offensive label}
At level A, we determine whether the comment is offensive (OFF) or not (NOT), and which part of the comment makes it offensive (offensive span). 
We consider a comment offensive if it contains any form of untargeted profanity or targeted offense such as insults and threats, which can be implicit or explicit~\citep{zampieri-etal-2019-predicting}.

We define offensive span as a specific segment of text that justifies why a comment is offensive, also known as a rationale \citep{zaidan2007using}. In parallel with the definition of offensiveness, the span includes not only explicit profanity but also implicit offensive language (e.g., sarcasm or metaphor \citep{elsherief-etal-2021-latent}). If the offensiveness is conveyed across multiple sentences in the comment, all of them are captured as the offensive span. Taking into account the \textit{faithfulness} of a rationale \citep{deyoung-etal-2020-eraser}, the offensive span is the minimal snippet of the text (i.e., sufficient) that includes all forms of expressions that convey even the slightest intensity of offense (i.e., comprehensive), such as affixes and emojis.


\subsection{Level B: Target Type Categorization of Offensive Language}
Level B categorizes the type of the target and highlights the supporting span of the target (target span).
There are four possible categories.
\begin{itemize}
    \item \textbf{Untargeted (UNT):} An offensive comment that does not contain a specific target. 
    \item \textbf{Individual (IND):} An offensive comment that is targeted at a specific individual. This includes a famous person or a named/unnamed individual with specific reference in the text. Comments targeted at an individual are categorized as cyberbullying \citep{chen2012detecting}.
    \item \textbf{Group (GRP):} An offensive comment targeted at a group of people with shared protected characteristics, such as gender or religion. Offensive language in this category is generally considered as hate speech~\citep{zhang2019hate}.
    \item \textbf{Others (OTH):} An offensive comment whose target does not belong to the above two categories. Targeting an organization, a company, or an event.
\end{itemize}

We define a target span as a span of characters in the comment that indicates the target of the offensive language.
It is collected for all types of targeted offensive speech, regardless of the target type (IND, GRP, OTH).
If the term used to indicate the target is offensive, target span can overlap with the offensive span (e.g., \textit{jjangkkae}, which corresponds to \textit{ching-chong} in English). 

\subsection{Level C: Target Group Identification of Group Targeted Offensive Language}

Level C identifies the specific targets of offensive language, which consists of two hierarchical levels: target group attribute and target group. 
The target group represents the specific social or demographic groups that share the same identity (e.g., Women, Muslim, Chinese), and the target group attribute is a superclass for the target group.
We allow multi-group annotation if the target entity of the comment belongs to more than one group. For instance, \textit{``페미년 (feminist bitch)''}, a word that disparages a feminist woman, targets two groups: \textit{Women} and \textit{Feminist}.
Table \ref{table:appendix_group} in the Appendix contains the full set of 21 target groups. To determine the set of target groups, we begin with categorizing targets in \citet{sap-etal-2020-social} and add several categories to better reflect the Korean language and culture.
As the result of analyzing the targets in 1,000 initial samples, we add \textit{Chinese}, \textit{Korean-Chinese}, and \textit{Indian} to the \textit{Race, Ethnicity \& Nationality} attribute, as they take up larger portions than the initial target groups (\textit{White}, \textit{Asian}). Group characteristics that do not belong to the four target group attributes (e.g., \textit{Disabled}, \textit{Feminist}) are grouped under \textit{Miscellaneous}. Note that we are aware that feminism is a gender-related issue, but classified \textit{Feminist} into \textit{Miscellaneous} because feminists embody a group of people that share the same ideology rather than being a subclass of gender.
We show the distribution of the top two levels (A, B) and the target group attributes (Level C) in Table~\ref{tab:distribution} and the subsequent target group categories in Table~\ref{tab:target_group_dist}.
\section{Collecting Annotations}

\subsection{Source Corpora Collection}

We choose two social media platforms, NAVER and YouTube, as our source of data, which are two of the top three mobile apps used in Korea in 2021.\footnote{\href{http://www.koreaherald.com/view.php?ud=20210901001000}{Using big data analysis to chart a new course, The Korea Herald}}
In particular, we collect titles and comments on NAVER news articles and YouTube videos distributed from March 2020 to March 2022. 

Due to the scarcity of offensive comments, we collect articles and posts by using predefined keywords, which is a commonly used method in hate speech dataset construction \citep{waseem2016hateful}. Every keyword is potentially highly correlated with articles or videos that may have abusive comments. Keywords are listed in Appendix \ref{appendix:a}.

To ensure we do not reveal users' personally identifiable information, we do not collect user ids. We replace mentions of a username with <user> tokens, URLs with <url> tokens, and emails with <email> tokens to conceal private information.

\subsection{Annotation Procedure}


The steps we took for high-quality of annotations include providing a detailed guideline, selecting the annotators deliberately, and managing the annotation process carefully. In the guideline, we resolve predictable difficulties during the process. For example, we provide rules for delimiting morphological boundaries specific to Korean to collect consistent text spans, and provide guidance of implicit hate speech based on the taxonomy proposed in \citet{elsherief-etal-2021-latent}. 
To ensure overall annotation quality, we only allow annotators who pass a qualification test to participate in the main annotation process. 
We make it clear that annotators refer to the title of the article or video as context to reduce ambiguity.
As hate speech annotation can also be influenced by the bias of the annotators \citep{davidson-etal-2019-racial, sap2021annotators, al-kuwatly-etal-2020-identifying}, we include annotators of diverse demographic backgrounds by limiting the maximum amount of annotation per worker to 1\% of the data. A total of 3,124 annotators participate in creating the final dataset.
To ensure each label is genuine, we embed questions with clear known answers and remove users who answer those wrong. We use SelectStar,
a crowdsourcing platform in Korea, to collect annotations. The full task is shown in the appendix (Figure \ref{fig:annotation}).


To decide on the gold label, we apply majority voting among the three annotations for the categorical labels, and take character offsets that more than two out of three annotators highlighted for text spans. When the gold label cannot be determined by majority voting as there are more than two choices, inspectors\footnote{Inspectors are selected workers who passed the qualification test with the highest scores.} resolve the disagreement to determine the gold label.  

\subsection{Annotation Result}

                     

\begin{table}[t!]
\centering
\resizebox{0.9\columnwidth}{!}{%
\begin{tabular}{@{}cccr@{}}
\toprule
\textbf{Level A} &
  \textbf{Level B} &
  \textbf{Level C} &
  \multicolumn{1}{c}{\textbf{Count (\%)}} \\ \midrule
NOT                  & -   & -        & 20,119 (49.8)  \\ \midrule
\multirow{13}{*}{OFF} & UNT & -        & 2,596 (6.4)   \\ \cmidrule(l){2-4} 
                     & IND & -        & 3,899  (9.6)   \\ \cmidrule(l){2-4} 
                     & OTH & -        & 1,402 (3.5)   \\ \cmidrule(l){2-4} 
                     
 &
  \multirow{8}{*}{GRP} &
  \begin{tabular}[c]{@{}c@{}}Gender \& \\ Sexual Orientation\end{tabular} &
  2,986 (7.4) \\ \cmidrule(l){3-4} 
 &
  &
  \begin{tabular}[c]{@{}c@{}}Race, Ethnicity \\ \& Nationality\end{tabular} &
  2,970 (7.3) \\ \cmidrule(l){3-4} 
                     &     & Political Affiliation & 1,939 (4.8)   \\ \cmidrule(l){3-4} 
                     &     & Religion & 1,699 (4.2) \\ \cmidrule(l){3-4} 
                     &     & Miscellaneous   & 2,819 (7.0) \\ \midrule
\multicolumn{3}{c}{Total}      & 40,429 (100) \\ \bottomrule
\end{tabular}%
}
\caption{Statistics of labels in KOLD. Here, the lowest level of data granularity is the target group attribute in Level C.}
\label{tab:distribution}
\end{table}

\begin{table}[!]
\centering
\resizebox{0.9\columnwidth}{!}{%
\begin{tabular}{@{}ccr@{}}
\toprule
\textbf{\begin{tabular}[c]{@{}c@{}}Target Group \\ Attribute\end{tabular}}                                                               & \textbf{Target Group}  & \textbf{Count} \\ \midrule
                                                                                        & LGBTQ+                & 1,369          \\
                                                                                        & Women                & 1,129          \\
                                                                                        & Men                  & 591            \\
\multirow{-4}{*}{\begin{tabular}[c]{@{}c@{}}Gender \& \\ Sexual Orientation\\ (24.63\%)\end{tabular}}    & Others                         & 108  \\ \midrule
                                                                                        & Chinese               & 475            \\
                                                                                        & Korean-Chinese        & 460            \\
                                                                                        & Black                 & 175            \\
\multirow{-4}{*}{\begin{tabular}[c]{@{}c@{}}Race, Ethnicity\\ \& Nationality\\ (23.93\%)\end{tabular}} & Others & 1,605 \\ \midrule
                                                                                        & Progressive           & 1,052          \\
                                                                                        & Conservative          & 316            \\
\multirow{-3}{*}{\begin{tabular}[c]{@{}c@{}}Political Affiliation\\ (15.55\%)\end{tabular}}          & Others                         & 651   \\ \midrule
                                                                                        & Muslim                & 1,059          \\
                                                                                        & Christian             & 518            \\
                                                                                        & Catholic              & 57             \\
\multirow{-4}{*}{\begin{tabular}[c]{@{}c@{}}Religion\\ (13.42\%)\end{tabular}} & Others            & 91          \\ \midrule
                                                                                        & Feminist             & 1,483               \\
                                                                                        & Socio-economic Status & 88                \\
                                                                                        & Agism                 & 62                \\
\multirow{-4}{*}{\begin{tabular}[c]{@{}c@{}}Miscellaneous\\ (20.32\%)\end{tabular}}    & Others            & 971          \\ \midrule
\multicolumn{2}{c}{Total}                                                                              & 12,981         \\ \bottomrule
\end{tabular}%
}
\caption{Breakdown of target group attributes of group-targeted offensive comments (Level C). We present the three most frequent target groups for each target group attribute. Multi-targeted groups are split into single groups when counting.}
\label{tab:target_group_dist}
\end{table}

Overall, the average Krippendorff's $\alpha$ for inter-annotator agreement of each annotation level is 0.55. The label distribution of the collected data is shown in Table \ref{tab:distribution}.

\paragraph{Level A: Offensive Language Detection}


The dataset contains 40,429 comments, of which 20,130 comments are classified as offensive language.
Offensive comments targeted at group characteristics (also known as hate speech) comprise 30.7\% of the whole data.
Specifically, Krippendorff's $\alpha$ for the inter-annotator agreement is 0.55 for classifying offensiveness.

\paragraph{Level B: Target Type of Categorization of Offensive Language}

Among 20,130 offensive comments, 2,596 comments are classified as untargeted offense (UNT). These comments include comments with non-targeted profanity and swearing. Group (GRP) is the most common type of target, taking up 70.1\% of the three types of targeted offenses, followed by individual (IND) (22.0\%) and others (OTH) (7.9\%).
Krippendorff's $\alpha$ for agreement on deciding the type of target is 0.45 .

\paragraph{Level C: Target Group Identification of Group Targeted Offensive Language}

Among 12,413 group-targeted offensive comments, the most common attribute is \textit{Gender \& Sexual orientation} taking up 7.4 of the whole dataset. \textit{Political Affiliation} and \textit{Religion} both appear in less than 5\% of the data. Table \ref{tab:target_group_dist} shows a breakdown of the target group attributes within the top three most frequent as well as the group \textit{Others}, which is tagged at the target group attribute level but does not belong to the target group choices we provide. In \textit{Race, Ethnicity \& Nationality}, \textit{Others} take up the largest portion with 1,605 comments, since it includes small but various origins ranging from Afghans and Americans to North Koreans and North Korean defectors. The three most frequently targeted group characteristics in the whole dataset are \textit{Feminist}, \textit{LGBTQ+}, and \textit{Women}, which amount to 11.42\%, 10.55\%, and 8.7\% of the group-targeted offensive language, respectively. 
Krippendorff's $\alpha$ is 0.65 for specifying the target group of the offensive language.

\section{Dataset Analysis}

\subsection{Target Group Distribution} 

Our novel finding is that target groups are defined based on the specific language and culture to embrace ongoing social phenomena and reflect them to the dataset. Shown in Table~\ref{tab:hatexplain}, the distribution of the target groups in KOLD largely differs from the English HateXplain dataset \citep{mathew2020hatexplain}. We observe that groups such as \textit{Jewish, Arabs} and \textit{Hispanic} which commonly appear in English datasets (e.g., \citet{sap-etal-2020-social, ousidhoum-etal-2019-multilingual}), do not frequently appear in KOLD. \textit{Africans}, the target group that appears most frequently in HateXplain, is not included in the top ten ranked groups in our dataset, and the reverse is true for \textit{Feminist}, the first-ranked target group in our dataset. While \textit{Women, LGBTQ+} (which includes Homosexual) and \textit{Men} are common target groups in both datasets, other identity groups such as \textit{Chinese, Korean-Chinese, Progressive} and \textit{Conservative} only appear in our dataset.

Specifically, we observe that within the Korean language, the Asian race as a target of offensive language should be more finely partitioned. 
While \textit{Asians} appear as a frequent target in English datasets without race or ethnic division \citep{an-etal-2021-predicting-anti, ousidhoum-etal-2019-multilingual, hartvigsen-etal-2022-toxigen}, in KOLD, it is further separated into fine-grained targets grouped by nationality or ethnicity (e.g., \textit{Chinese}, \textit{Indian}, \textit{Southeast Asian}). Moreover, our dataset demonstrates that a single Asian race should be divided into separate target groups. A large portion of \textit{Chinese} and \textit{Korean-Chinese} targeted offensive comments in KOLD highlights the uniqueness of offensive language in Korean and reflects the cultural differences between Korean speakers and English speakers. This demonstrates the prevalence of social bias among Asians even though they share similar cultural values and phenotype \citep{lee2017neo}.

\begin{table}[!]
\resizebox{\columnwidth}{!}{%
\begin{tabular}{@{}ccccc@{}}
\toprule
\multirow{2}{*}{\textbf{Rank}} & \multicolumn{2}{c}{\textbf{Ours}}          & \multicolumn{2}{c}{\textbf{HateXplain}}     \\ \cmidrule(l){2-3} \cmidrule(l){4-5} 
                               & \textbf{Target group} & \textbf{\%} & \textbf{Target group} & \textbf{\%} \\ \midrule
1  & \colorbox[HTML]{EBC0A9}{Feminist}       & 11.42 & \colorbox[HTML]{B5EBE9}{African}    & 21.58 \\
2  & LGBTQ+         & 10.55 & \colorbox[HTML]{B5EBE9}{Jewish}     & 13.04 \\
3  & Women         & 8.7  & Muslim     & 12.4  \\
4  & Muslim          & 8.16   & Homosexual & 10.22 \\
5  & \colorbox[HTML]{EBC0A9}{Progressive}    & 8.1   & Women      & 9.5   \\
6  & Men            & 4.55  & \colorbox[HTML]{B5EBE9}{Arab}       & 6.17  \\
7  & \colorbox[HTML]{EBC0A9}{Christian}      & 3.99  & \colorbox[HTML]{B5EBE9}{Refugee}    & 3.46  \\
8  & \colorbox[HTML]{EBC0A9}{Chinese}        & 3.66  & \colorbox[HTML]{B5EBE9}{Caucasian}  & 3.11  \\
9  & \colorbox[HTML]{EBC0A9}{Korean-Chinese} & 3.54  & \colorbox[HTML]{B5EBE9}{Hispanic}   & 2.74  \\
10 & \colorbox[HTML]{EBC0A9}{Conservative}   & 2.43  & Men        & 2.13  \\ \bottomrule
\end{tabular}%
}
\caption{Comparison of target group distribution with existing dataset in English (HateXplain). We list the top 10 target groups of each dataset based on the frequency.\footnotemark~Groups in \colorbox[HTML]{EBC0A9}{brown} only appear in KOLD, while groups in \colorbox[HTML]{B5EBE9}{mint} only appear in HateXplain.}
\label{tab:hatexplain}
\end{table}
\footnotetext{We calculate the statistics of the HateXplain dataset ourselves as \citet{mathew2020hatexplain} do not provide the specific distribution of the target groups. We changed the target group \textit{Islam} to \textit{Muslim}, to match with our term.}

\subsection{The Role of Title for Target Group Identification}

In KOLD, 55.1\% of group-identified offensive comments have no target span marked in the comment, which implies that in the majority of the cases, titles contain information about the targets. For example, given the title of the article \textit{```Islam' in Korea / Yonhap News''}, it is easy to find a comment without explicitly mentioning the target such as \textit{``I don't care what (they) believe, what matters is the fact that (they) kill people''} (penultimate row of Table~\ref{tab:my-table}).

\section{Experiments and Results}

We experiment using three different model architectures: (1) sequence classification model for predicting offensiveness, target type, and target group categories, (2) token classification model for predicting the offensive and target span, and (3) multi-task model for predicting both category and span at once. 
We report the score of single-task models using various sizes of Korean BERT and RoBERTa \cite{park2021klue}, and compare the result against the multi-task model.

We further conduct an ablation study by excluding the title from the input to discover how much it impacts the prediction performance of the model. We also compare the results of our model with translated versions of English data, and a multilingual span prediction model.

For the experiments, we use an 80-10-10 split for each task, and report the best performances based on the $F_1$ score of the test set result with the tuned hyperparameters. Training details are reported in the Appendix \ref{appendix:b}.

\subsection{Category Prediction}
\label{sec:category}
\begin{table}[t]
\centering
\resizebox{0.95\columnwidth}{!}{
\begin{tabular}{cclccc}
\toprule
                                     & \multicolumn{2}{c}{}                               & \multicolumn{3}{c}{\textbf{Metric}}                                                                 \\ \cmidrule(l){4-6} 
\multirow{-2}{*}{}   & \multicolumn{2}{c}{\multirow{-2}{*}{\textbf{Model}}} & \textbf{P}           & \textbf{R}              & \textbf{F1}                           \\ \midrule
                                     & \multicolumn{2}{c}{\fontfamily{qcr}\selectfont BERT$_{base}$}                 & 75.8                         & 78.8                         & 77.2                                  \\
                                     & \multicolumn{2}{c}{\fontfamily{qcr}\selectfont RoBERTa$_{base}$}              & 76.8                         & 77.6                         & 77.2                                  \\
\multirow{-3}{*}{\textbf{\begin{tabular}[c]{@{}c@{}}Level A: \\ Offensive\end{tabular}}} & \multicolumn{2}{c}{\fontfamily{qcr}\selectfont RoBERTa$_{large}$}             & 76.6                        & 81.1                         & \textbf{78.8} \\ \midrule
                                     & \multicolumn{2}{c}{\fontfamily{qcr}\selectfont BERT$_{base}$}                 & 58.5                         & 58.5                         & 57.6                                  \\
                                     & \multicolumn{2}{c}{\fontfamily{qcr}\selectfont RoBERTa$_{base}$}              & 60.8                         & 58.7                         & 59.4                                  \\
\multirow{-3}{*}{\textbf{\begin{tabular}[c]{@{}c@{}}Level B: \\ Target Type\end{tabular}}}    & \multicolumn{2}{c}{\fontfamily{qcr}\selectfont RoBERTa$_{large}$}             & 63.9                         &62.4                         & \textbf{62.7} \\ \midrule
                                     & \multicolumn{2}{c}{\fontfamily{qcr}\selectfont BERT$_{base}$}                 & 56.9                         & 57.4                         & 55.9                                  \\
                                     & \multicolumn{2}{c}{\fontfamily{qcr}\selectfont RoBERTa$_{base}$}              & 58.6                         & 58.5                         & 57.3                                  \\
\multirow{-3}{*}{\textbf{\begin{tabular}[c]{@{}c@{}}Level C: \\ Target Group\end{tabular}}}     & \multicolumn{2}{c}{\fontfamily{qcr}\selectfont RoBERTa$_{large}$}             & 59.1                         & 59.6                         & \textbf{57.5} \\ \bottomrule
\end{tabular}
}
\caption{Evaluation results of category prediction models of each task. Binary-F1 is used to evaluate \textit{offensive} task and macro-F1 is used for the other tasks whose number of labels is more than two. \textbf{Bold} indicates the best performance across the models.}
\label{tab:sentence_classification}
\end{table}

For each level of annotation, we fine-tune the pre-trained models to predict the label given the title and comment, and then evaluate the model using precision, recall, and $F\textsubscript{1}$ scores of the positive class.

In Table~\ref{tab:sentence_classification}, we observe that the more the categories are fine-grained, the task becomes more difficult, and the larger model shows better performance.
\begin{table}[t]
\centering
\resizebox{0.9\columnwidth}{!}{
\begin{tabular}{cclccc}
\toprule
                                                                                                   
\multirow{2}{*}{}   & \multicolumn{2}{c}{\multirow{2}{*}{\textbf{Model}}} & \multicolumn{3}{c}{\textbf{Metric}}               \\ \cmidrule(l){4-6}
                                    & \multicolumn{2}{c}{}                              & \textbf{P}    & \textbf{R}    & \textbf{F1}       \\ \midrule
\multirow{3}{*}{\textbf{\begin{tabular}[c]{@{}c@{}}Offensive \\ Span\end{tabular}}}                                     & \multicolumn{2}{c}{\fontfamily{qcr}\selectfont BERT$_{base}$}                 & 47.7               & 43.4            & 41.9          \\
                                     & \multicolumn{2}{c}{\fontfamily{qcr}\selectfont RoBERTa$_{base}$}              & 46.4               & 41.8            & 40.6          \\
                                     & \multicolumn{2}{c}{\fontfamily{qcr}\selectfont RoBERTa$_{large}$}             & 50.8               & 47.8            & \textbf{45.4} \\ \midrule
\multirow{3}{*}{\textbf{\begin{tabular}[c]{@{}c@{}}Target \\ Span\end{tabular}}}                                         & \multicolumn{2}{c}{\fontfamily{qcr}\selectfont BERT$_{base}$}                 & 60.2               & 67.3            & 61.9          \\
                                     & \multicolumn{2}{c}{\fontfamily{qcr}\selectfont RoBERTa$_{base}$}              & 60.6               & 68.2            & 62.4          \\
& \multicolumn{2}{c}{\fontfamily{qcr}\selectfont RoBERTa$_{large}$}            & 60.8               & 67.8            & \textbf{62.5} \\ \bottomrule
\end{tabular}
}
\caption{Evaluation results of offensive/target span prediction models. \textbf{Bold} indicates the best performance across the models.}
\label{tab:span_prediction}
\end{table}
\begin{table*}[h!]
\centering
\small
\begin{tabular}{@{}c|cccccc|cccccc@{}}
\toprule
\multirow{2}{*}{\textbf{System}} & \multicolumn{3}{c}{\textbf{Offensive}} & \multicolumn{3}{c|}{\textbf{Offensive Span}} & \multicolumn{3}{c}{\textbf{Target Type}} & \multicolumn{3}{c}{\textbf{Target Span}} \\ \cmidrule(l){2-4} \cmidrule(l){5-7} \cmidrule(l){8-10} \cmidrule(l){11-13} 
                                 & \textbf{P}     & \textbf{R}    & \textbf{F1}    & \textbf{P}    & \textbf{R}   & \textbf{F1}   & \textbf{P}     & \textbf{R}     & \textbf{F1}     & \textbf{P}  & \textbf{R}  & \textbf{F1}  \\ \midrule
Seq                              & 75.8           & 78.7          & \textbf{77.2}           & -    & -   & -    & 58.5           & 58.5           & \textbf{57.6}            & -  & -  & -   \\
Span                             & -     & -    & -     & 47.7          & 43.4         & 41.9          & -     & -     & -      & 60.2        & 67.3        & 61.9         \\
Seq + Span                       & 75.0           & 78.1          & 76.5           & 56.3          & 55.4         & \textbf{51.7}          & 56.8          & 57.0          & 56.5           & 72.3        & 71.9        & \textbf{71.6}       \\
\bottomrule
\end{tabular}
\caption{Evaluation results of the multi-task baseline models compared against the single-task baseline models.  "Seq" refers to the sequence classification model, "Span" refers to the span prediction model, and "Seq+Span" refers to the multi-task model. All models have fine-tuned the same pre-trained language model, BERT-base.
}
\label{tab:multitask}
\end{table*}

\subsection{Span Prediction}
We convert each span in the comment to BIO-tags to formulate the span prediction task as a token classification task and fine-tune the pre-trained models to predict BIO-tags assigned to each token.
To evaluate the model, we follow the work of \citet{da-san-martino-etal-2019-fine} by computing the $F_1$ score of the predicted character offsets with the ground truth. If the ground truth is empty, a perfect score ($F_1 = 1$) is assigned whereas if the predicted set of offsets is empty, a score of zero ($F_1 = 0$) is assigned. 

As demonstrated in Table \ref{tab:span_prediction}, the best character-level F1 score is 45.4 for offensive span and 62.5 for target span. 
The pattern of higher score with a larger model is consistent with the results of the category prediction.

\subsection{Category and Span Prediction}

We employ a multi-task learning approach to train a model capable of classifying the category and predicting the span at the same time.  
In the multi-task model, the sequence classifier and the token classifier share the neural representation of the pre-trained model and only differ in the output layers for each task. The representation of the first token ([CLS]) is fed into an output layer for sequence classification, and the other representations are fed into the layer for token classification. The model jointly learns the global information of a given input sequence and span information.
We train two types of multi-task models. 
First, we train a multi-task model with the binary label of offensiveness and the corresponding offensive span. Second, using the data labeled as offensive, we train a multi-task model to predict the target type (Level B) and the corresponding target span.

As shown in Table \ref{tab:multitask}, multi-task models outperform single-task models in span prediction by 10\% in the F1 score, mainly due to joint learning of both types of information. 
However, the performance of sequence classification drops in both models.

Examples of model predictions and ground truths are illustrated in Table~\ref{tab:example_multitask_a} and Table~\ref{tab:example_multitask_b} in the Appendix.

\begin{table}[t]
\centering
\resizebox{\columnwidth}{!}{
\begin{tabular}{cccc}
\toprule
\multicolumn{1}{c}{} \textbf{Level} & \textbf{Comment} & \textbf{Comment+Title} & \textbf{Δ F1} \\ \midrule
\textbf{A}   & 76.9             & 77.2                   & + 0.3 (0.4\%)        \\
\textbf{B}      & 56.1             & 57.6                   & + 1.5 (2.6\%)        \\
\textbf{C}       & 42.8             & 55.9                   & + 13.1 (30.6\%)      \\ \bottomrule
\end{tabular}
}
\caption{Category prediction results on the offensiveness (Level A), target type (Level B), and target group categories (Level C) with and without the title. Binary-F1 is used to evaluate the offensive task and macro-F1 is used for the other tasks. ΔF1 refers to the absolute and relative performance gap. For all models, we use the same BERT-base model.}
\label{tab:title_ablation}
\end{table}
\subsection{Title Ablation on Classification Tasks}
\label{sec:ablation}
To find out how much the context information (titles of the articles and the videos) contributes to the offensiveness and target classifications, we conduct an ablation study by excluding the titles from the input.

As can be seen in Table~\ref{tab:title_ablation}, if only the comments are given, accuracy and the f1 scores drop significantly compared to the setting where titles and comments are given together.
This phenomenon becomes more significant as the granularity of the label increases.
When predicting the fine-grained target group, the f1 score dropped by more than 30\%. 
We conclude that providing the context with the comments helps the model predict the target groups more precisely, as the comments may not contain sufficient information. 

\subsection{Translated Data and Multilingual Models}
\begin{table}[!t]
\centering
\small
\begin{tabular}{clc}
\toprule
                                                                           & \textbf{System}  & \textbf{F1}   \\ \midrule
\multirow{2}{*}{\textbf{Offensive}}                                                 & KOLD\textsubscript{Seq}        & \textbf{76.9} \\
                                                                                    & OLID\textsubscript{EN→KO}    & 63.2          \\ \midrule
\multirow{2}{*}{\textbf{\begin{tabular}[c]{@{}c@{}}Offensive \\ Span\end{tabular}}} & KOLD\textsubscript{Span}       & \textbf{40.6} \\
                                                                                    & MUDES            & 12.8         \\ \bottomrule
\end{tabular}
\caption{OLID\textsubscript{EN→KO} is an offensiveness classification model trained with the English OLID dataset translated into Korean. MUDES is a multilingual offensive span prediction model. We compare the results against our baseline performance.}
\label{tab:trans}
\end{table}

To see how much translation and multilingual model are effective at distinguishing offensiveness, we compare our baselines against (1) sequence classification model trained on translated dataset, and (2) multilingual offensive span detection model.
For the translation experiment, we translate the OLID to Korean via google translate api\footnote{Requested in April 2022.} and use the dataset for training the same sequence classifier described in Section~\ref{sec:category}. For the multilingual experiment, we adopt the multilingual token classification model (MUDES)~\citep{ranasinghe-zampieri-2021-mudes} trained on English toxic span dataset \citep{pavlopoulos-etal-2021-semeval}. For all tasks, evaluation is done with the KOLD test set. We report the results in Table \ref{tab:trans}.

Overall, both translation and multilingual approaches are not more effective than our baselines. For the offensive category prediction, our model is 13.7 higher, and for the span prediction, our model is 27.8 higher. Although MUDES scores high on English (61.6), the performance drops significantly in Korean (12.8). 
\section{Related Work}

\subsection{Offensiveness \& Hate Speech Detection}
Most datasets created for the detection of offensive language have dealt with the subtypes of offensive language such as hate speech, cyberbullying, and profanity as a flat multi-level classification task \citep{waseem2016hateful, davidson2017automated, wiegand2018overview, mollas2022ethos}.
\citet{waseem2017understanding} and \citet{zampieri-etal-2019-predicting} have proposed a hierarchical taxonomy of offensive speech, emphasizing the need for annotating specific dimensions of offensive language, such as the content's explicitness and the type of targets. \citet{rosenthal-etal-2021-solid} further expands the size of the dataset using the OLID dataset proposed by \citet{zampieri-etal-2019-predicting} with semi-supervising method. The hierarchical annotation has also made possible systematic expansion to subtypes of hate speech in the following works, such as misogyny \citep{zeinert-etal-2021-annotating}. Our work also builds upon the taxonomy proposed by \citet{zampieri-etal-2019-predicting}, further identifying the targeted social group of offensive languages.

Recent papers focus on more diverse aspects, such as interpretability and context information. To train a human-interpretable classification models, \citet{sap-etal-2020-social} collect social bias implicated about the targeted group in a free-text format. In a similar spirit, \citet{pavlopoulos-etal-2022-detection} and \citet{mathew2020hatexplain} create datasets annotated with particular span of the text that makes the post toxic \citep{zaidan2007using}.
As most text in the real world appears in context \citep{seaver2015nice}, considering context is important for the development of practical models. Recent work on offensive language detection incorporates the context of the post \citep{vidgen-etal-2021-introducing,de-gibert-etal-2018-hate,gao-huang-2017-detecting}, albeit the benefits of the context are controversial \citep{pavlopoulos-etal-2020-toxicity, xenos-etal-2021-context}. 
Using hierarchical annotation, KOLD dataset systematically classifies multiple depths of contextualized offensiveness, and collects textual spans to justify such classification at the same time.

\subsection{Non-English Datasets}
There is relatively little work done on developing offensive language datasets in languages other than English.  
Simple translation of English datasets is not enough as there are some well-known issues in using automatically translated English datasets in NLP, such as \textit{translationese}~\citep{koppel2011translationese} and over-representation of source language's culture \citep{hu-etal-2020-ocnli}. Several papers have emphasized the need of high-quality monolingual data \citep{hu-etal-2021-investigating, park2021klue}. This is also true in offensive language datasets. The focus of hatred differs by culture and country \citep{reichelmann-2021-hate-knows}. \citet{ousidhoum-etal-2019-multilingual} observe that there are significant differences in terms of target attributes and target groups in the three languages (English, French, Arabic) of which they constructed hate speech datasets. Moreover, \citet{nozza2021exposing} shows that zero-shot, cross-lingual transfer learning of English hate speech has limitations.
Some datasets for detection of toxicity or abuse exist in other languages (e.g., \citet{zeinert-etal-2021-annotating} for Danish, \citet{fortuna-etal-2019-hierarchically} for Portuguese, \citet{mubarak-etal-2021-arabic} for Arabic, and \citet{ccoltekin2020corpus} for Turkish). For Korean, \citet{moon-etal-2020-beep} have paved the way for hate speech detection, but they are relatively small in size and lack focus on the target of offensiveness.
In comparison, KOLD is built upon an extensive taxonomy that can handle a broad range of offensive language with clearly annotated target groups of 21 categories and textual spans.
\section{Conclusion}
We present KOLD, a dataset of 40,429 comments of news articles and video clips, annotated within context. It is the first to introduce a hierarchical taxonomy of offensive language in Korean with textual spans of the offensiveness and the targets.
We establish baseline performance for multi-task model that both detects the categories and the spans that support the classification. 
Through analysis and experiments, we show that target terms are often omitted in offensive comments, and title information helps models predict the target of the offense. This finding can be applied to other syntactically null-subject languages other than Korean (e.g., Arabic, Chinese, Modern Greek) as well. 
By comparing the distribution of target groups with existing English data and showing the inadequacy of multilingual models, we demonstrate that offensive language corpus customized for the language and its corresponding culture is necessary. We acknowledge that our dataset does not cover all communities of Korean social media whose offensive language patterns may differ from each other. Despite this limitation, KOLD will serve as a stepping stone to developing more accurate and adaptive offensive language detection models in Korean. 
\section{Ethical Considerations}
This study has been approved by the KAIST Institutional Review Board (\#KH2021-177).
During the annotation process, we informed the annotators that the content might be offensive or upsetting and limited the amount that each worker could provide.
Annotators were also paid above the minimum wage.
We are aware of the risk of releasing a dataset containing offensive language. This dataset must not be used as training data to automatically generate and publish offensive language online, but by publicly releasing it, we cannot prevent all malicious use. We will explicitly state that we do not condone any malicious use. We urge researchers and practitioners to use it in beneficial ways (e.g., to filter out hate speech). Another consideration is that the names of political figures and popular entertainers mentioned in the comments remain in our dataset. This is because offensive language detection becomes difficult without those mentions. This is consistent with the common practice in other offensive language datasets, and as a community, we need to deliberate and discuss the potential implications.
\section{Limitations}
We discuss two limitations of our work in this section. First, our annotation method requires a high annotation cost and a lot of time. Guiding the annotators to familiarize them with our annotation process takes much time since the guideline is complicated, and they are updated whenever ambiguous comments are reported during the process, to give annotators clear direction.
Furthermore, as we collect three annotations per comment, we need a large number of annotators (3,124) and spent a significant amount of annotation cost to pay them above the minimum wage. 

In most cases, there is a trade-off between quantity and quality. For example, if one plans to build a large-scale dataset with limited amount of resources, he/she should sacrifice the complexity of the annotations by reducing the amount of work for each annotator, or the accuracy of the labels by decreasing the number of annotators for each sample.
This is the reason why it is challenging to build a large dataset with accurate and rich annotations. Recently, there has been an approach to make a large-scale machine-generated hate speech detection dataset \cite{hartvigsen-etal-2022-toxigen}. This might be an alternative to overcome such limitations. By collaborating with such models, we can obtain large-scale datasets with accurate labels while reducing annotation costs and time. 

Second, detecting patterns of offensive language changing over time requires constant update. For example, hateful comments related to COVID-19 emerged recently, and offensive language toward political figures or celebrities also changes constantly. It is difficult to train a model that captures such changes well with a dataset within a limited time period. A model trained on our dataset might not perform so well in detecting hateful comments that emerge in the future. To overcome this limitation, a continuous update of datasets as well as methods to efficiently update models \cite{qian-etal-2021-lifelong}, is needed. 
\section*{Acknowledgments}
This project was funded by the KAIST-NAVER Hypercreative AI Center. Alice Oh is funded by Institute of Information \& communications Technology Planning \& Evaluation (IITP) grant funded by the Korea government (MSIT) (No. 2022-0-00184, Development and Study of AI Technologies to Inexpensively Conform to Evolving Policy on Ethics). SelectStar provided a crowdsourcing platform for the annotation of the data.
\bibliographystyle{acl_natbib}
\bibliography{anthology,custom}
\clearpage
\section*{Appendix}
\appendix
\section{Data Collection Keywords}
\label{appendix:a}


\begin{itemize}
    \item Gender : 백래시 (backlash), 여성 단체 (female organization), 여성 혐오 (misogyny), 젠더 (gender), 페미니즘 (feminism)
    \item Sexual orientation : 동성혼 (homosexual marriage), 성소수자 (sexual minority), 차별금지법 (anti-discrimination legislation), 퀴어 (queer), 퀴어 활동가 (queer activist)
    \item Race \& Ethnicity \& Nationality : 난민 (refugee), 동남아 (Southeast Asia), 백인 (White), 외국인 근로자 (foreign worker), 이민자 (immigrant), 인도 (India), 조선족 (Korean-Chinese), 중국 동포 (ethnic Korean from China), 탈북민 (North Korean defectors), 흑인 (Black)
    \item Religion : 이슬람 (Islam), 탈레반 (Taliban), 기독교 (Christianity), 카톨릭 (Catholic), 교회 (Church), 목사 (minister)
\end{itemize}

\begin{table*}[t]
\centering
\resizebox{\textwidth}{!}{
\begin{tabular}{ll}
\toprule
\textbf{Target Group Attribute}           &   \textbf{Target Group}                                                    \\ \midrule
Gender \& Sexual Orientation        &   LGBTQ+, Men, Women                                                      \\
Race \& Ethnicity \& Nationality    &   Asian, Black, Chinese, Indian, Korean-Chinese, Southeast Asian, White   \\
Political Affiliation               &   Conservative, Progressive                                               \\
Religion                            &   Buddhism, Catholic, Christian, Islam                                    \\
Miscellaneous                       &   Agism, Disabled, Diseased, Feminist, Physical Appearance, Socio-economic Status \\ \bottomrule
\end{tabular}
}
\caption{Target group attributes and target groups in the KOLD dataset.}
\label{table:appendix_group}
\end{table*}

\begin{figure*}[t]
    \centering
    \includegraphics[width=\textwidth]{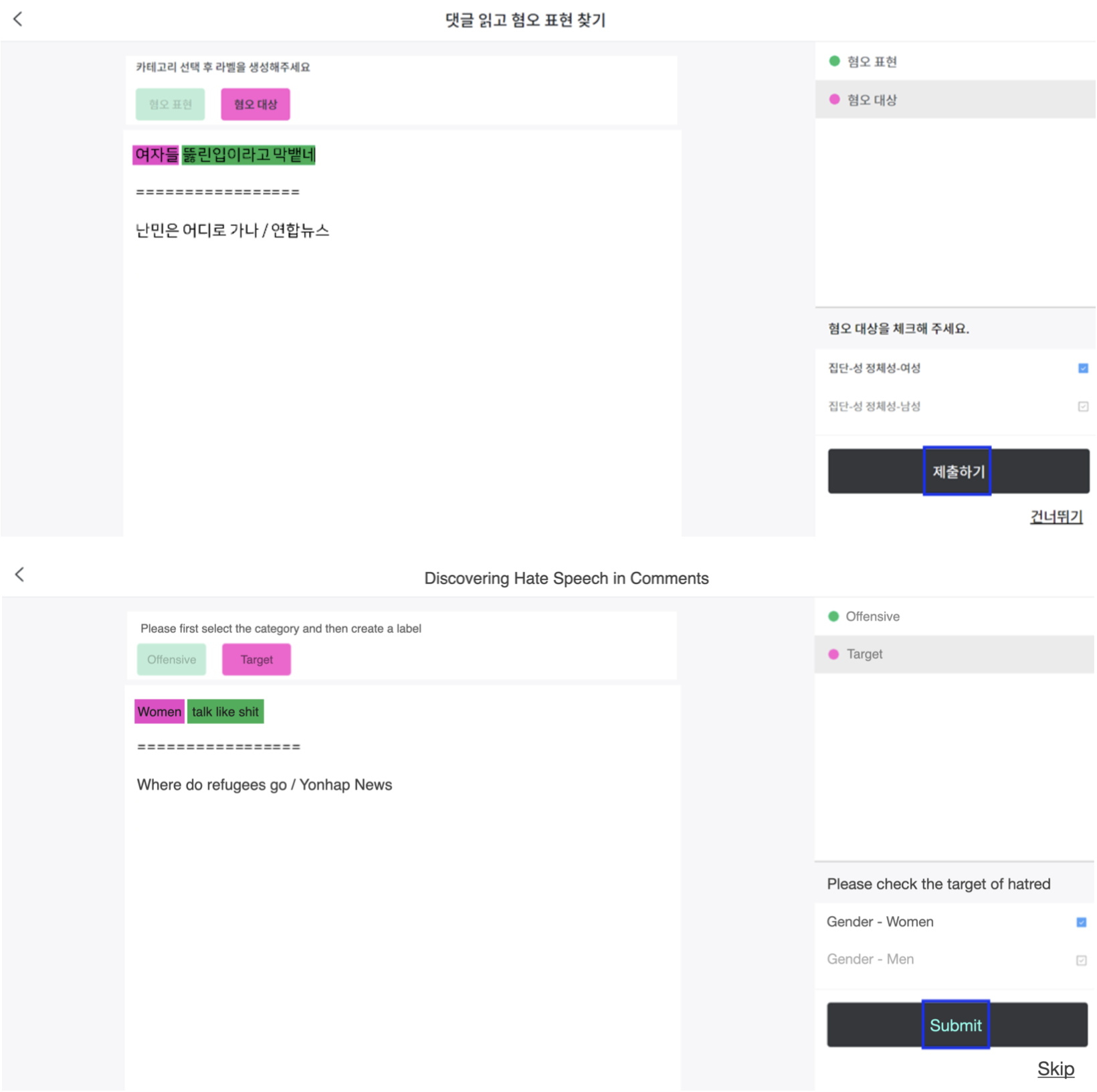}
    \caption{Annotation interface used to collect label and span annotations of KOLD.}
    \label{fig:annotation}
\end{figure*}

\section{Training Details}
\label{appendix:b}
All experiments are conducted using the Transformers library via HuggingFace\footnote{\href{https://github.com/huggingface/transformers}{https://github.com/huggingface/transformers}}. 
For the experiments, we searched for a learning rate out of \{1e-5, 2e-5, 3e-5, 5e-5\} and the number of epochs out of \{1, 2, 3, 4, 5\}. We kept the batch size fixed at 32 for training, 64 for validation. We report a mean score of 5 runs. The experiments are conducted on GeForce RTX 2080 Ti 10GB, with 10.2 CUDA version. The single experiment takes from an hour to 4 hours. The number of parameters for BERT base model is 110 million, 123 million for RoBERTa base and 354 million for RoBERTa large.

\begin{table*}[!t]
\centering
\resizebox{\textwidth}{!}{
\begin{tabular}{ccll}
\toprule
Gold & Pred & \multicolumn{1}{c}{Title}           & \multicolumn{1}{c}{Comment}                   \\ \midrule
    OFF        & OFF        & \begin{tabular}[c]{@{}l@{}}미국서 백인 여성 한국계 부부에 욕설 \\ White female curses to Korean-American couples in the U.S \end{tabular}               & \begin{tabular}[c]{@{}l@{}} 미국은 소수의 똑똑한사람들이 전체를 먹여살리는거임 \\ 나머지는 걍 \colorbox[HTML]{C1D6EB}{\textbf{\underline{개돼지 아이큐수준}}} \\ In the US, few elite people feed the whole country \\ The rests are just \colorbox[HTML]{C1D6EB}{\textbf{\underline{the great unwashed with low-level IQ}}} \end{tabular} \\ \midrule
OFF        & OFF        & \begin{tabular}[c]{@{}l@{}}차별금지법을 반대하는 교회들에게 \\ To churches who oppose anti-discrimination laws  \end{tabular}                  & \begin{tabular}[c]{@{}l@{}} 요즘 \colorbox[HTML]{C1D6EB}{\textbf{\underline{개독들이 신천보다 더함 개독은 종교도 않임}}} \\ These days \colorbox[HTML]{C1D6EB}{\textbf{\underline{Jesus freaks are way worse than Shincheonji.}}} \\ \colorbox[HTML]{C1D6EB}{\textbf{\underline{Halle-fxxkin-lujah is not a even religion.}}} \end{tabular}                                 \\ \midrule
NOT        & NOT        & \begin{tabular}[c]{@{}l@{}} 캐나다 이민 | 다시 한국으로 돌아가고 싶어요. \\ Immigration to Canada | I want to go back to Korea. \end{tabular}          & \begin{tabular}[c]{@{}l@{}} 자연의 위치..저수지 에 사는.물고기가 바다에.살면.. \\ 적응하기 어려운. 환경에.. \\ The law of nature... When a fish living in reservoirs goes to the sea \\ it has to have many difficulties to adapt  \end{tabular} \\ \midrule[1.9pt]

OFF        & \colorbox[HTML]{FCA087}{NOT}        & \begin{tabular}[c]{@{}l@{}} 탈동성애자가 밝히는 동성애의 실체 \\ The reality of homosexuality revealed by ex-gay \end{tabular}              & \begin{tabular}[c]{@{}l@{}} 난 \colorbox[HTML]{C1D6EB}{이 재미난 동성애에 빠져버렸어 $\sim$$\sim$$\sim$} \\ I'm \colorbox[HTML]{C1D6EB}{into this funny homosexuality $\sim$$\sim$$\sim$} \end{tabular}               \\ \midrule
OFF        & \colorbox[HTML]{FCA087}{NOT}        & \begin{tabular}[c]{@{}l@{}} 이준석, 차별금지법, 기독교 \\ Lee Jun-seok, Anti-discrimination law, Christianity \end{tabular}   & \begin{tabular}[c]{@{}l@{}} 이준석 ... 차별금지법을 옹호한다고 ??? \\ \colorbox[HTML]{C1D6EB}{이준석은 \textbf{\underline{OUT}}} \\ 
Lee Jun-seok ... advocate the anti-discrimination law ??? \\ \colorbox[HTML]{C1D6EB}{Lee Jun-seok is \textbf{\underline{OUT}}} \end{tabular}   \\ \midrule
OFF        & \colorbox[HTML]{FCA087}{NOT}        & \begin{tabular}[c]{@{}l@{}} 미국이 만든 난민은 스스로 책임져야 \\ US has to take responsibility for the refugees they created \end{tabular}                     & \begin{tabular}[c]{@{}l@{}} \colorbox[HTML]{C1D6EB}{어중떠중천하} \\ \colorbox[HTML]{C1D6EB}{A bunch of ragtag} \end{tabular}  \\
\bottomrule
\end{tabular}
}
\caption{Examples of Level A (Offensiveness) predictions for labels and spans by a multi-task model. Top 3 examples show that the multi-task model predicts correct labels and spans, however, below 3 examples show its errors in labels and spans. (OFF: offensive, NOT: not offensive, \colorbox[HTML]{C1D6EB}{blue}: gold span, \textbf{\underline{underline}}: predicted span, \colorbox[HTML]{FCA087}{red}: incorrect label)}
\label{tab:example_multitask_a}
\end{table*}

\begin{table*}[!t]
\centering
\resizebox{\textwidth}{!}{%
\begin{tabular}{ccll}
\toprule
Gold & Pred & \multicolumn{1}{c}{Title}           & \multicolumn{1}{c}{Comment}                   \\ \midrule
GRP & GRP   & \begin{tabular}[c]{@{}l@{}} 조선족조폭의 강남 장악기 가리봉 잔혹사 \\ Korean-Chinese Gang's Domination in Gangnam \\ the cruel history of Garibong \end{tabular}   &   \begin{tabular}[c]{@{}l@{}} 정부가 무능해서 그렇지  한심한 \colorbox[HTML]{C9DFB8}{\textbf{\underline{정치인}}} \\ Because the government is incompetent. Pathetic \colorbox[HTML]{C9DFB8}{\textbf{\underline{politician}}} \end{tabular} \\ \midrule
GRP & GRP   & \begin{tabular}[c]{@{}l@{}} {[}정치와 젠더 세미나 2회차{]} '젠더와 군대' \\ {[}Politics and Gender Seminar 2nd {]} 'Gender and the Army' \end{tabular} &  \begin{tabular}[c]{@{}l@{}} \colorbox[HTML]{C9DFB8}{\textbf{\underline{정의당}}}에 정의는 어디? \\ Where is justice in \colorbox[HTML]{C9DFB8}{\textbf{\underline{Justice Party}}}? \end{tabular} \\ \midrule
IND & IND   & \begin{tabular}[c]{@{}l@{}} "미국으로 가자!" 과테말라 국경 뚫은 이민자들 \\ "Let's go to America!" Immigrants Breaking the Guatemala Border \end{tabular} & \begin{tabular}[c]{@{}l@{}} 또라이 \colorbox[HTML]{C9DFB8}{\textbf{\underline{바이든}}} \\ Freak \colorbox[HTML]{C9DFB8}{\textbf{\underline{Biden}}} \end{tabular} \\ \midrule[1.9pt]

GRP & GRP   &  \begin{tabular}[c]{@{}l@{}} "코로나보다 무서운 건 혐오의 시선"... 코로나 시대의 중국동포 \\ "Eyes of Hatred scarier than COVID-19"... \\ The age of COVID-19, Compatriots in China \end{tabular}  & \begin{tabular}[c]{@{}l@{}} \underline{\textbf{동포}}는 뭔 개소리냐 \\ \underline{Compatriots}? What the h**l are you talking about? \end{tabular} \\ \midrule
GRP & GRP   & \begin{tabular}[c]{@{}l@{}} {[}한국교회 CPR{]} 삯꾼 목사 청소는 얼마면 돼요? \\ {[}Korean Church CPR{]} How much is it to clean up \\ all the wage-earner pastors? \end{tabular} & \begin{tabular}[c]{@{}l@{}} \colorbox[HTML]{C9DFB8}{\textbf{\underline{맘}}몬}이 강단을 장악했네ㅜㅜ \\ 합동을 떠나야 하나..안에서 싸워야 하나 ㅜㅜ \\ \colorbox[HTML]{C9DFB8}{\textbf{\underline{Mum}}-Roach} took control of podium ;( \\ Should I leave the general meeting or fight against ;( \end{tabular}   \\ \midrule
GRP & \colorbox[HTML]{FCA087}{UNT}   &   \begin{tabular}[c]{@{}l@{}} 탈영병, 중국서 조선족 4명 살해 \\ Deserters Killed 4 Korean-Chinese in China \end{tabular} &  \begin{tabular}[c]{@{}l@{}} 착한 살인 인정합니다 \\ I admit that it's a good murder \end{tabular} \\
\bottomrule
\end{tabular}%
}
\caption{Examples of Level B (Target) predictions for labels and spans by a multi-task model. Top 3 examples show that the multi-task model predicts correct labels and spans, however, below 3 examples show its errors in labels and spans. (UNT: untargeted, IND: individual, GRP: group, OTH: other, \colorbox[HTML]{C9DFB8}{green}: gold span, \textbf{\underline{underline}}: predicted span, \colorbox[HTML]{FCA087}{red}: incorrect label)}
\label{tab:example_multitask_b}
\end{table*}

\end{document}